\documentclass{article}
\usepackage{natbib}
\usepackage[utf8]{inputenc}
\usepackage{url}
\usepackage{hyperref}
\usepackage{multirow}
\usepackage{array}
\usepackage{graphicx}
\usepackage{xcolor}
\usepackage{marvosym}
\usepackage{amssymb}
\usepackage{authblk}
\usepackage{hyperref}
\newcommand*{\email}[1]{%
    {\small \texttt{#1}}\par
}
\providecommand{\keywords}[1]
{
  \small	
  \textbf{\textit{Keywords---}} #1
}

\title{The Ecological Footprint of Neural Machine Translation Systems}

\author{Dimitar Shterionov\footnote{Corresponding author}~, Eva Vanmassenhove\\
Department of Cognitive Science\\
And Artificial Intelligence\\
Tilburg University \\
\email{\{d.shterionov, e.o.j.vanmassenhove\}@tilburguniversity.edu}}

\date{March 2021}

\begin{document}
\maketitle

\begin{abstract}
Over the past decade, deep learning (DL) has led to significant advancements in various fields of artificial intelligence, including machine translation (MT). These advancements would not be possible without the ever-growing volumes of data and the hardware that allows large DL models to be trained efficiently. Due to the large amount of computing cores as well as dedicated memory, graphics processing units (GPUs) are a more effective hardware solution for training and inference with DL models than central processing units (CPUs). However, the former is very power demanding. The electrical power consumption has economical as well as ecological implications. 

This chapter focuses on the ecological footprint of neural MT systems. It starts from the power drain during the training of and the inference with neural MT models and moves towards the environment impact, in terms of carbon dioxide emissions. Different architectures (RNN and Transformer) and different GPUs (consumer-grate NVidia 1080Ti and workstation-grade NVidia P100) are compared. Then, the overall CO2 offload is calculated for Ireland and the Netherlands. The NMT models and their ecological impact are compared to common household appliances to draw a more clear picture.

The last part of this chapter analyses quantization, a technique for reducing the size and complexity of models, as a way to reduce power consumption. As quantized models can run on CPUs, they present a power-efficient inference solution without depending on a GPU.
\end{abstract}

\section{Introduction}
Over the past decade, Deep Learning (DL) techniques took the world by storm and their application led to state-of-the-art results in various fields. The same holds for the field of Machine Translation (MT), where the latest AI boom led the way for the emergence of a new paradigm \emph{Neural Machine Translation} (NMT). Since most of the foundational techniques used in current applications of AI were developed before the turn of the century, the main triggers of the boom were innovations in general purpose GPU computing as well as hardware advancements that facilitate much more efficient training and inference. In spite of the improvements in terms of efficiency, GPUs are more power demanding than their CPU predecessors and as such, they have a considerably higher environmental impact.

In this chapter, a brief overview of the most recent paradigms is presented along with a more in depth introduction to the core processing technology involved in NMT (GPUs as opposed to CPUs). We elaborate on the exponential growth of models, the trend of `big data' and their relation to model performance. The related work discusses pioneering and recent papers on Green AI for Natural Language Processing (NLP) as well as tools to quantify the environmental impact of AI. 


As a case study and to outline the realistic dimensions of power consumption and environmental footprint of NMT, we train 16 NMT models and use them for translation, while collecting power readings from the corresponding GPUs. Using the collected measurements we compare different NMT architectures (Long Short-Term Memory (LSTM) and Transformer), different GPUs (NVidia GTX 1080Ti and NVidia Tesla P100) as well as the environmental footprint of training and translation with these models to other, commonly-used devices. 

Together with the research and analytical contributions, this chapter also aims to motivate researchers to devote time, efforts and investments in developing more ecological solutions for MT.

\section{The Technological Shift(s) in MT}
Machine translation (MT), the task of automatically translating text in one language, into text in another language using a computer system, has become an indispensable tool for professional translators (to assist in the translation workflow), for commercial users, e.g. e-commerce companies (to make their content quickly available in multiple languages), to every-day users (to access information unrestricted by the language in which it is produced). Since its inception in the late 1950s, MT has undergone many shifts, the latest of which, neural machine translation (NMT) imposes a requirement for hardware that can facilitate efficient training and inference with NMT models. Most commonly used hardware are graphics processing units (GPUs) that support embarrassingly parallel computations due to the large amounts of processing cores and dedicated memory.\footnote{There are other hardware and software that are specifically developed for AI accelerated computing, e.g. tensor processing units (https://cloud.google.com/tpu/docs/tpus). However, as the most commonly used and easily accessible such devices are GPUs, our work focuses on the power considerations and environmental footprint of GPUs.} 

\subsection{From rule-based to neural MT}
In the early days of machine translation (MT), rule-based MT (RBMT) systems were built around dictionaries and human-crafted rules to convert a source sentence into its equivalent in the target language. Such systems were heavily dependant on the efforts and skills of linguists. Developing a rule-based system for a new domain or a new language pair was (an still is) a cumbersome, time-consuming task that requires extensive linguistic expertise~\citep{arnold2001introductory}. However, from a computational point of view, it is an inexpensive task -- using an RBMT system does not require substantial computational resources. 

In the 1980s, researchers attempted overcoming some of the shortcomings of RBMT when dealing with languages that differed substantially structurally (e.g. English vs Japanese). Focusing specifically on collocations, examples could be used for transfer when rules and trees failed~\citep{Nagao1984_ExampleBasedMT}. These hybrid approaches evolved by the end of the decade into a more example-centered approach (example-based MT (EBMT)), where patterns would be retrieved from existing corpora and adapted using hand-written rules~\citep{hutchins2005towards}. The idea of using patterns extracted from corpora culminated when, in the early 90s, a group of researchers at IBM created an MT system relying solely on statistical information extracted from corpora~\citep{hutchins2005history}.


The generation of corpus-based MT systems rely on data and statistical models to derive word-, phrase- or segment-level translations eliminating the need for complex semantic and/or syntactic rules. As such, the core mechanism of MT systems shifted from human expertise in linguistics to machine learning techniques. This shift entailed other important changes for MT related to the development time and the computational resources required for training\footnote{As a matter of fact, the technical advances in terms of computational resources facilitated developments in the field of corpus-based MT. The paradigm shift furthermore coincided with the late 90s and early 2000s growth in terms of direct applications for MT and localisation.}. Furthermore, this group of MT paradigms that learn automatic translation models from large amounts of parallel and monolingual data reached -- for in-domain translations and given enough available training data -- a better overall translation quality than that achieved by earlier RBMT systems.

Up until about 2016, Phrase-Based Statistical MT (PB-SMT) was the dominant corpus-based paradigm (especially after Google Translate made the switch from RBMT to SMT)~\citep{Bentivogli2016}. Currently, most state-of-the-art results for MT are achieved using neural approaches, i.e. models based on artificial neural networks and most prominently recurrent neural networks (RNNs)~\citep{Sutskever2014,bahdanau2014neural,Cho2014} and Transformer architectures~\citep{Vaswani2017}. RNNs, as the name suggests, feed their output as input, along with new input. This enables RNNs to compress sequences (of tokens, e.g. words) of arbitrary length into a fixed-size representation. This representation can then be used to initiate a decoding process where one token is generated at a time (again using a recurrent network) conditioned on the previously generated tokens and the encoded representation of the input until a certain condition is met. In the context of MT, this generation process typically continues until the end-of-sentence token is generated. To mitigate issues related to long-distance dependencies, LSTM units~\citep{Zaremba2014} or gated recurrent units~\citep{Cho2014} are typically used instead of simple RNNs. In addition, to further improve the relation between encoder and decoder, an attention mechanism is added~\citep{Luong2015_attention_NMT}. The attention mechanism learns to associate different weights based on the importance of individual input tokens. It has been shown that attention-based models significantly outperform those that do not employ attention. In contrast to NMT using RNNs, Transformer does not employ recurrence; it uses self attention, i.e. an attention mechanism that indicates the importance of a token with respect to the other input tokens. Within a self-attention mechanism the positional information is lost. As such, Transformer employs a positional encoding that learns the positional information. These operations can be performed in parallel for different tokens, which allows the training process of a Transformer model to be parallelised, unlike RNNs, where operations are performed sequentially, making Transformer a more efficient architecture.

For a more complete overview and description of the history of Machine Translation, we refer to the work of Hutchins~\citeyearpar{hutchins2005towards}, Kenny~\citeyearpar{kenny2005parallel}, Poibeau~\citeyearpar{poibeau2017machine} and Koehn~\citeyearpar{koehn2020neural}.

\subsection{Why GPUs?}
\begin{figure}[h]
    \centering
    \includegraphics[width=0.75\textwidth]{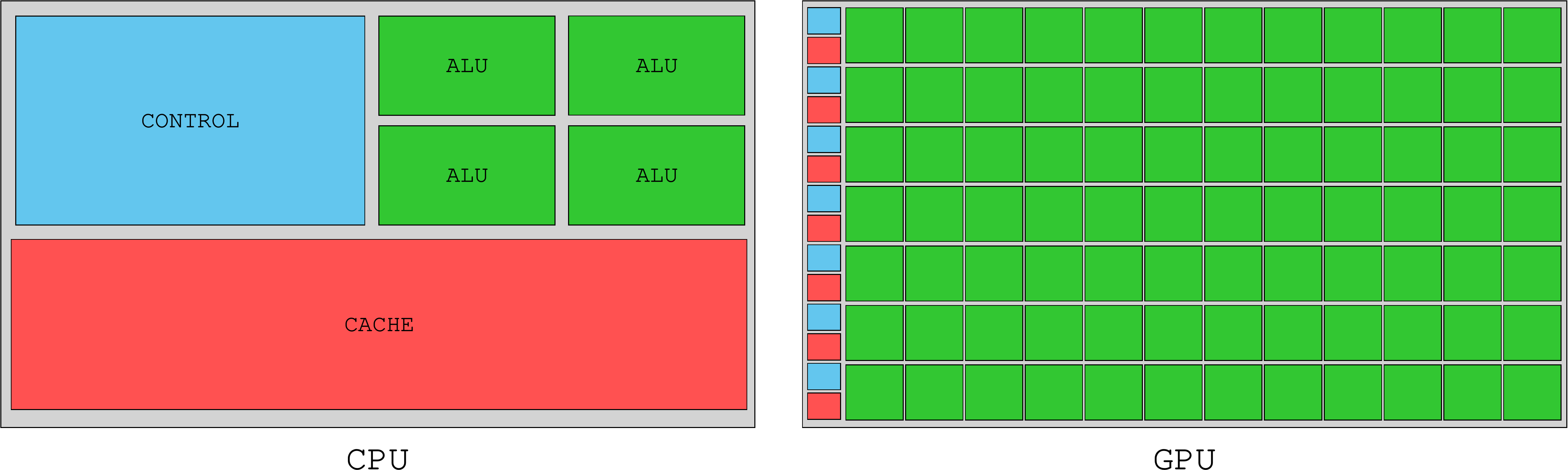}
    \caption{Visual comparison between a CPU and a GPU in terms of cache, control and processing units. Source: \url{https://fabiensanglard.net/cuda/}}
    \label{fig:cpu_vs_gpu}
\end{figure}

Neural MT (NMT) evolved very quickly, replacing PB-SMT in both academia and industry. Aside from the paradigm shift, NMT imposed another big change within the field of MT -- that of the core processing technology. In particular, employing graphics processing units (GPUs) instead of central processing units (CPUs). Training neural networks (NNs) revolves around the manipulation of large matrices. A general-purpose, high-performance CPU typically contains 16 to 32, high-frequency cores that can operate in parallel. CPUs are designed for general purpose, sequential operations. However, the sizes of matrices involved in training NNs are far beyond the processing and memory capacities of CPUs; processes that can (theoretically) be parallelised need to be serialized and conducted in a sequence, leading to rather high processing times. To perform all matrix-operations efficiently a parallel-processing framework is more suitable. Graphics processing units (GPUs) are designed to render and update graphics. They encapsulate thousands of cores with tens of thousands of threads. With their large dedicated memory GPUs can host both an NN model and training examples reducing the required memory transfer. The large amount of processing cores that can operate in parallel and the dedicated memory makes GPUs a much more effective option for training NN models~\citep{Raina2009}, including NMT models. See Figure~\ref{fig:cpu_vs_gpu} for a visual comparison between a CPU and a GPU in terms of cache, control and processing units. 

In GPUs, however, many more transistors are dedicated to data processing, rather than to caching and control flow as in CPUs~\citep{Raina2009}. GPUs are in fact much more power demanding than CPUs leading to two types of considerations: (i) first, the physical power requirements towards a data center or a workstation dedicated to training NN models, and (ii) second, the ecological concern related to the production and consumption of electricity to power up and sustain the training of NN, including NMT, models. In this chapter we are going to focus on the second question. We will present actual power and thermal indicators measured during the training of different NMT models and align them with ecological as well as economical markers in Section~\ref{sec:power}. Then, in Section~\ref{sec:mitigation}, we will discuss two approaches to reduce the GPU power consumption -- distribution and parallelisation, and quantization.

\subsection{The more, the better?}

Since statistical models (both traditional and deep learning) took over in the field of Natural Language Processing (NLP), datasets and models grew bigger. Especially since 2018 when BERT~\citep{Devlin2019BERTPO} and its successors (a.o. GPT 2~\citep{radford2019language}, GPT-3~\citep{brown2020language} and Turing NLG~\citep{microsoft_2020}) appeared, the size of language models and the amount of parameters grew exponentially. Since the relation between a model's performance and its complexity is at best logarithmic~\citep{Schwartz2020GreenA}, exponentially larger models are being trained for often small gains in performance. This exponential growth is illustrated well by the Switch-C, the current largest language model introduced in 2021 with a capacity of 1.6 trillion parameters. For comparison, one of its recent predecessors GPT-3,  currently the third\footnote{The second largest model is the GShard~\citep{lepikhin2020gshard}, introduced in September 2020 which had a capacity of 600 billian parameters.} largest model, introduced in June 2020 had a capacity of `only' 175 billion parameters.\footnote{\url{https://analyticsindiamag.com/open-ai-gpt-3-language-model/}} Similarly, a blogpost by OpenAI~\citep{open_ai_2018} demonstrated how the compute grew by more than 300.000x. This corresponds to a doubling every 3.4 months~\footnote{For comparison, Moore's Law forecasted a doubling every 2 years for the number of transistors in a dense integrated circuit~\citep{open_ai_2018}}

Many studies have investigated the performance of NMT in terms of adequacy, fluency, errors, data requirements, impact on the translation workflow and language service providers, bias. The efficiency and energy impact of developing new Neural MT (NMT), however, has not yet received the necessary attention. In Section~\ref{sec:background} we present related work in order to properly position our research in current literature. 

\section{Related work}\label{sec:background}

The related work is divided into three subsections: Section~\ref{subsec:rel_paper}, Section~\ref{subsec:rel_tools} and Section~\ref{subsec:rel_workshops}. In Section~\ref{subsec:rel_paper}, we cover the related research papers. Since there are a few more practical tools that have been suggested and constructed in order to measure the environmental and financial cost, we cover the main tools in Section~\ref{subsec:rel_tools}. Finally, in Section~\ref{subsec:rel_workshops}, we mention some recent initiatives related to sustainable NLP.

\subsection{Research}\label{subsec:rel_paper}
Recent work~\citep{strubell-etal-2019-energy, Schwartz2020GreenA} brought to the attention of the NLP community the environmental (carbon footprint) and financial (hardware and electricity or cloud compute time) cost of training `deep' NLP models. In Strubel et al.~\citeyearpar{strubell-etal-2019-energy} the energy consumption (in kilowatts) of different state-of-the-art NLP models is estimated. With this information, the carbon emissions and electricity costs of the models can be approximated. From their experiments regarding the cost of training, it results that training BERT~\citep{Devlin2019BERTPO} is comparable to a trans-American flight in terms of carbon emissions. They also quantified the cost of development by studying logs of a multi-task NLP model~\citep{strubell-etal-2018-linguistically} that received the Best Long Paper award at EMNLP 2018. From the estimated development costs, it resulted that the most problematic aspect in terms of cost is the tuning process and the full development cycle (due to hyperparameter grid searches)\footnote{During tuning the model is trained from an already existing checkpoint, typically using new data.}$^,$\footnote{In the development cycle, different versions of the model are trained or tuned and evaluated. Each of those differ in terms of hyperparameter values.} and not the training process of a single model. They conclude their work with three recommendations for NLP research which stress the importance of: (i) reporting the time required for (re)training and the hyperparameters' sensitivity, (ii) the need for equitable access to computational resources in academia, and (iii) the development of efficient hardware and models. 

Similar to \cite{strubell-etal-2019-energy}, the work by Schwartz et al.~\citeyearpar{Schwartz2020GreenA} advocates for `Green AI', which is defined as ``AI research that is more environmentally friendly and inclusive'' \citep{Schwartz2020GreenA} and is directly opposed to environmentally unfriendly, expensive and thus exclusive `Red AI'. Although the `Red AI' trend has led to significant improvements for a variety of AI tasks, Schwartz et al.~\citeyearpar{Schwartz2020GreenA} stress that there should also be room for other types of contributions that are greener, less expensive and that allow young researcher and undergraduates to experiment, research and have the ability to publish high-quality work at top conferences. The trend of so-called `Red AI', where massive models are trained using huge amount of resources, can almost be seen as a type of `buying' stronger results, especially given that the relation between the complexity of a model and its performance is at best logarithmic implying that exponentially larger model are required for linear gains. Nevertheless, their analysis of the trends in AI based on papers from top conferences such as ACL\footnote{\url{https://acl2018.org}} and NeurIPS~\footnote{\url{https://nips.cc/Conferences/2018}} reveals that there is a strong tendency within the field to focus merely on the accuracy (or performance) of the proposed models with very few papers even mentioning other measures such as speed, model size or efficiency. They propose making the efficiency of models a key criterion aside (or integrated with) commonly used metrics. There are multiple ways to measure efficiency\footnote{For a more detailed overview of measures we refer to the paper itself.}, floating point operations (FPO) being the one advocated for by Schwartz et al.~\citeyearpar{Schwartz2020GreenA}. FPO is a metric that has occasionally been used to determine the energy footprint of models~\citep{molchanov2016pruning,Vaswani2017,gordon2018morphnet,veniat2018learning} as it estimates ``the work performed by a computation process''~\citep{Schwartz2020GreenA} based on two operations `ADD' and `MUL'. They furthermore advocate for reporting a baseline that promotes data-efficient approaches by plotting accuracy as a function of ``computational cost and of training set size''~\citep{Schwartz2020GreenA}. Aside from the environmental impact, both recent papers stress also the importance of making research more inclusive and accessible and thus advise reviewers of journals and conferences to recognize contributions to the field by valuing certain contributions (e.g. based on efficiency) even when they do not beat state-of-the-art results.

While in the context of a different topic, that of NLP leaderboards, in its critique on how the community's focus has shifted towards accuracy rather than applicability, the work of \cite{Ethayarajh2020_NLPleaderboards_critique} points out the importance of efficiency, fairness, robustness and other desiderata on applicability of NLP models in practice. In fact, in a commercial setting where revenue is the main criterion of success, efficiency (i.e. of training / updating models, or of inference) is often more important than accuracy. See, for example, the work of~\cite{shterionov-etal-2019-less} which compares several systems for quality estimation of MT based on performance metrics, business metrics and efficiency (training and inference times). This publication stems from an academic-industry collaboration, based on commercial data and use-cases.

Another recent, yet heavily debated\footnote{Among others due to the fact that it has been considered the centerpiece that led to Google firing leading A.I. ethics researcher Dr. Timnit Gebru, and subsequently Dr. Margaret Mitchell, two of the authors of the paper.}, paper by \cite{bender2021dangers}, focuses on a broader range of problems related to current trends in A.I. and the risks associated with these trends. Their paper goes beyond merely discussing the environmental impact of large models as it covers various concerns related specifically to the unfathomable nature of large data sets and its potential consequences (related to e.g. bias, derogatory associations, stereotyping), including for downstream tasks. The section that focuses specifically on environmental and financial cost mainly summarizes findings of previous papers (e.g. \cite{strubell-etal-2019-energy} and \cite{Schwartz2020GreenA}) as well as some recent tools and benchmarks. They also provide a table with an overview of the 12 most recent, largest language models along with their number of parameters and the data set sizes, which illustrates well how these massive language models have grown exponentially over the last two years. The paper furthermore points out that the majority of these technologies are constructed for the already privileged part of society while marginalized populations are the ones more likely to experience environmental racism~\citep{barsh1990indigenous,Pulido2016flint,bender2021dangers}. 

Both \cite{lacoste2019quantifying} and \cite{henderson2020towards} provide tools (see Section~\ref{subsec:rel_tools} for more details) to help researchers report carbon emissions and energy consumption. Aside from the tools themselves, both papers also provide an explanation of certain factors (such as server location, energy grid, hardware and the run-time of training procedures) and their impact on the emissions as well as more practical mitigation guidelines for researchers. The recommendations and best practices listed in \cite{lacoste2019quantifying} center around transparency, the choice of cloud providers, the location of the data center, a reduction of wasted resources and the choice of hardware. \cite{henderson2020towards} provides additional mitigation and reporting strategies highlighting how both industry and academia can improve their environmental impact in terms of carbon and energy efficiency. Their mitigation strategies are an extension of those provided in \cite{lacoste2019quantifying} and include: ensuring the reproducibility of experiments (and thus avoiding replication difficulties), focusing more on energy-efficiency and energy-performance trade-offs, reducing overheads for utilizing efficient algorithms and selecting efficient test environments.

\subsection{Tools}~\label{subsec:rel_tools}
Quantifying the environmental impact of AI technology is a difficult task that might have prevented researchers from reporting energy consumption and/or carbon emissions. To overcome this problem and to make this information more easy to report and calculate, \cite{lacoste2019quantifying} released an online tool.\footnote{\url{https://mlco2.github.io/impact}} Given the run-time, hardware and cloud provider, the emission calculator estimates the raw carbon emissions and the approximate offset carbon emissions of your research. The cloud provider is of importance since the emissions incurred during training depend on the energy grid and the location of the training server. The main goal of the tool is to push researchers for transparency within the field of Machine Learning by publishing the `Machine Learning Emissions Calculator' results in their publications.

More recently, and similar to \cite{lacoste2019quantifying},~\cite{henderson2020towards} hypothesized that the complexity of collecting and estimating energy and carbon metrics has been one of the main bottlenecks for researchers. To encourage researchers to include such metrics in their work, they present the `experiment-impact-tracker'.\footnote{\url{https://github.com/Breakend/experiment-impact-tracker}} Their framework aims to facilitate ``consistent, easy and more accurate reporting of energy, compute and carbon impacts of ML systems''~\citep{henderson2020towards}. The authors claim that the assumptions made by previous estimation methods \citep{lacoste2019quantifying,Schwartz2020GreenA} lead to to significant inaccuracies, particularly for experiments relying heavily on both GPUs and CPUs.

\subsection{Workshop}~\label{subsec:rel_workshops}
\textbf{SustaiNLP}\footnote{https://sites.google.com/view/sustainlp2020}:  
The first SustaiNLP workshop was held in November, 2020 and (virtually) co-located with EMNLP2020.\footnote{https://2020.emnlp.org/} It specifically focused on efficiency, by encouraging researchers to design solutions that are more simple yet competitive with the state-of-the-art, and justifiability, by stimulating researchers to provide justifications for models and in this way encouraging for more novel and creative designs.

\section{Case Study: Empirical evaluation of MT systems}
In the following three sections we will try to give a realistic image of the power consumption and environmental footprint, in terms of carbon emissions, related to the usage of GPUs for training and translating with NMT models. To do that, we train multiple NMT models, using both LSTM and Transformer architectures on different GPUs. We record the power consumption for each GPU during training as well as during translation. In Section~\ref{sec:power} we analyse the results and in Section~\ref{sec:mitigation} we present one possible strategy to reduce power consumption at inference time, i.e. quantization -- the process of approximating a neural network's parameters by reducing their precision, thus reducing the size of a model.

\subsection{Hardware setup}
To assess the energy consumption during the training and translation processes of NMT we trained different NMT models from scratch, i.e. without any pretraining nor relying on additional models (e.g. BERT) on two different workstations, equipped with different GPUs. The GPUs we had at our disposal are four NVidia GeForce 1080Ti with 11GB of vRAM and 3 NVidia Tesla P100 with 16GB or vRAM. These units differ not only in terms of technical specifications, but also in their purpose of use -- the 1080Ti is a user-class GPU, developed for desktop machine with active cooling; the P100 is designed for workstations that operate continuously and does not support active cooling. 

In Table~\ref{tbl:gpu_differences}, we compare the two GPU types in terms of their specifications.

\begin{table}
    \centering
    \begin{tabular}{|r|c|c|}\hline
        & \textbf{1080Ti} & \textbf{P100} \\\hline
CUDA cores & 3584 & 3584\\\hline
vRAM & 11 GB & 16 GB\\\hline
Core clock speed & 1481 MHz & 1190 MHz\\\hline
Boost clock & 1600 MHz & 1329 MHz\\\hline
Transistor count & 11,800 million & 15,300 million\\\hline
Manufacturing process technology & 16 nm & 16 nm\\\hline
Power consumption (TDP) & 250 Watt & 250 Watt\\\hline
Maximum GPU temperature & 91 $^{\circ}$C & 85 $^{\circ}$C \\\hline
Floating-point performance & 11,340 gflops & 10,609 gflops\\\hline
Type & Desktop & Workstation\\\hline
    \end{tabular}
    \caption{An overview of the GPU specifications of 1080Ti and P100.}
    \label{tbl:gpu_differences}
\end{table}

The rest of the configurations are as follows: (i) for the 1080Ti desktop workstation -- an Intel(R) Core(TM) i7-7820X CPU @ 3.60GHz, RAM: 64GB (512MB block) (ii) for the P100 workstation -- CPU: Intel(R) Xeon(R) Gold 6128 CPU @ 3.40GHz, RAM: 196 GB (1GB block).

In this work we focus on assessing the power consumption related to the utilization of GPUs. As such, in this chapter we will not provide metrics related to the CPU utilization and the corresponding power consumption. This decision is motivated by the fact that GPUs are the main processing unit for NMT models.

\subsection{Machine Translation Systems}\label{sec:mt}
We experimented with the current state-of-the-art data-driven paradigms, RNN (LSTM) and Transformer. We used data from the Europarl corpus~\citep{Koehn2005} for two language pairs, English--French and English--Spanish in both direction (EN$\rightarrow$FR, FR$\rightarrow$EN, EN$\rightarrow$ES and ES$\rightarrow$EN). 
Our data is summarised in Table~\ref{tbl:data_stats}. For reproducibility reasons, the model specifications, data, scripts and our results are made publicly available for comparison on \url{https://github.com/dimitarsh1/NMT-EcoFootprint.git}.

\begin{table}[htb]
\centering
{\small
    \begin{tabular}{|c|c|c|c|}\hline
        Lang. pair & Train & Test & Dev \\\hline
        EN-FR/FR-EN & 1,467,489 & 499,487 & 7,723\\
        EN-ES/ES-EN & 1,472,203 & 459,633 & 5,734\\\hline
    \end{tabular}}
    \caption{Number of parallel sentences for the training, testing and development sets.}
    \label{tbl:data_stats}
\end{table}
We ought to note that our test set is atypically large. One of the main reasons for these experiments is that it can give us measures over a larger period of time enabling us to draw more stable conclusions. 

For the RNN and Transformer systems we used OpenNMT-py.\footnote{\url{https://opennmt.net/OpenNMT-py/}} The systems were trained for maximum 150K steps, saving an intermediate model every 5000 steps or until reaching convergence according to an early stopping criteria of no improvements of the perplexity (scored on the development set) for 5 intermediate models. The options we used for the neural systems are: 
\begin{itemize}
    \item RNN: size: 512, RNN type: bidirectional LSTM, number of layers of the encoder and of the decoder: 4, attention type: MLP, dropout: 0.2, batch size: 128, learning optimizer: Adam~\citep{Kingma2014} and learning rate: 0.0001.
    \item Transformer: number of layers: 6, size: 512, transformer\_ff: 2048, number of heads: 8, dropout: 0.1, batch size: 4096, batch type: tokens, learning optimizer Adam with beta$_2 = 0.998$, learning rate: 2.
\end{itemize}
All NMT systems have the learning rate decay enabled and their training is distributed over 4 nVidia 1080Ti GPUs. The selected settings for the RNN systems are optimal according to~\cite{Britz2017}; for the Transformer we use the settings suggested by the OpenNMT community\footnote{\url{http://opennmt.net/OpenNMT-py/FAQ.html}} as the optimal ones that lead to quality on par with the original Transformer work~\citep{Vaswani2017}. 

For training, testing and validation of the systems we used the same data. To build the vocabularies for the NMT systems we used sub-word units, allowing NMT to be more creative. Using sub-word units also mitigates to a certain extent the out of vocabulary problem. To compute the sub-word units we used BPE with 50,000 merging operations for all our data sets. Separate subword vocabularies were used for every language.
In Table~\ref{tbl:data_voc} we present the vocabulary sizes of the data used to train our PB-SMT and NMT systems.
\begin{table}[ht]
\centering
{\small
    \begin{tabular}{|c|c|c|c|c|}\hline
         & \multicolumn{2}{c|}{no BPE} & \multicolumn{2}{c|}{with BPE} \\\cline{2-5}
        Lang. pair & EN & FR/ES & EN & FR/ES \\\hline
        EN-FR/FR-EN & 113,132 & 131,104 & 47,628 & 48,459\\
        EN-ES/ES-EN & 113,692 & 168,195 & 47,639 & 49,283\\\hline
    \end{tabular}}
    \caption{Vocabulary sizes. For completeness we also present the vocabulary size without BPE, i.e. the number of unique words in the corpora.}
    \label{tbl:data_voc}
\end{table}

The quality of our MT systems is evaluated on the test set using standard evaluation metrics -- BLEU~\citep{Papineni2002} (as implemented in SacreBLEU~\citep{post-2018-call} and TER~\citep{Snover2006} (as implemented in MultEval~\citep{clark-etal-2011-better}). Our evaluation scores are presented in Table~\ref{tbl:mt_eval}. 

\begin{table}[thb]
\centering
{\small \setlength\tabcolsep{4pt}
    \begin{tabular}{|c|l|c|c|c|c||c|c|c|c|}
    \multicolumn{2}{c}{} & \multicolumn{4}{c}{English as source} & \multicolumn{4}{c}{English as target}\\\hline
    & & \multicolumn{2}{c|}{EN$\rightarrow$FR} & \multicolumn{2}{c||}{EN$\rightarrow$ES} & \multicolumn{2}{c|}{FR$\rightarrow$EN} & \multicolumn{2}{c|}{ES$\rightarrow$EN}\\\cline{3-10}
    & System  & BLEU$\uparrow$ & TER$\downarrow$ & BLEU$\uparrow$ & TER$\downarrow$& BLEU$\uparrow$ & TER$\downarrow$ & BLEU$\uparrow$ & TER$\downarrow$ \\\hline
\multirow{2}{*}{\rotatebox[origin=c]{90}{{\small 1080Ti}}} & LSTM & 34.2 & 50.9 & 38.2 & 45.3 & 34.6 & 48.2 & 38.1 & 44.7\\\cline{2-10}
& TRANS & \textbf{37.2} & \textbf{48.7} & \textbf{40.9} & \textbf{43.4} & \textbf{37.0} & \textbf{46.4} & \textbf{41.3} & \textbf{41.4}\\\hline\hline
\multirow{2}{*}{\rotatebox[origin=c]{90}{P100}} & LSTM & 34.1 & 50.7 & 37.3 & 47 & 34.9 & 48 & 38.5 & 44.4\\\cline{2-10}
& TRANS & \textbf{37.4} & \textbf{48.4} & \textbf{40.9} & \textbf{43.3} & \textbf{37.3} & \textbf{47} & \textbf{41.6} & \textbf{42.5}\\\hline
    \end{tabular}
    }
    \caption{Quality evaluation scores for our MT systems. TRANS denotes Transformer systems.}
    \label{tbl:mt_eval}
\end{table}
We computed pairwise statistical significance using bootstrap resampling~\citep{koehn-2004-statistical} and a 95\% confidence interval. The results shown in Table~\ref{tbl:mt_eval} are all statistically significant based on 1000 iterations and samples of 100 sentences. All metrics show the same performance trends for all language pairs: Transformer (TRANS) outperforms all other systems, followed by PB-SMT, and LSTM.

\subsection{GPU power consumption}
To measure the consumption of power, memory and core utilisation, as well as the heat generated during the training and use of an NMT model, several tools can be exploited, e.g. \verb|experiment-impact-tracker|\footnote{https://github.com/Breakend/experiment-impact-tracker}\citep{henderson2020towards}, ``Weights and Biases''\footnote{https://wandb.ai}, as well as NVidia's device monitoring: \verb|nvidia-smi dmon|. We decided to stick to the mainstream NVIDIA System Management Interface program (\verb|nvidia-smi|)~\footnote{\url{https://developer.download.nvidia.com/compute/DCGM/docs/nvidia-smi-367.38.pdf}} which does not require additional installation or setup. It is a tool by NVidia for monitoring and management of major lines of their GPUs. The \verb|nvidia-smi| is cross-platform and supports all standard NVidia driver-supported Linux distributions as well as 64bit versions of the Windows operating system. In this work we used the \verb|nvidia-smi dmon| command to monitor all GPUs during the training and inference processes. This command displays one line of monitoring data per monitoring cycle. The default range of metrics includes power usage (or power draw -- the last measured power draw for the entire board reported in watts), temperature, SM clocks, memory clocks and utilization values for SM, memory, encoder and decoder. 
By default we monitor all GPUs during the training process as training is distributed over all four 1080Ti or three P100 GPUs. However, at inference time we use only one GPU. As such, we only monitor and report values for that specific GPU during inference. 

We ought to note that the \verb|experiment-impact-tracker|, which internally invokes \verb|nvidia-smi|, is a much more elaborate tool, designed to ease the collection and analysis of data. However, it utilizes Intel's RAPL interface.\footnote{Intel's Running Average Power Limit or \emph{RAPL}\citep{Intel2009RAPL} interface exposes power meters and power limits. It also allows power limits on the CPU and the DRAM to be set.} As part of the Linux kernel, RAPL is read/write protected to normal users and the information it generates is readable by super users only. As such, the \verb|experiment-impact-tracker|, which utilizes RAPL would not collect any metrics from the CPU. We did not find any work-around but to disable this functionality. That, in turn, resulted in only collecting metrics from \verb|nvidia-smi| and therefore we did not employ the tool. However, along with the resource monitoring, this tool can compute the CO2 emissions based on the compute time and the energy consumed. 

The CO2 emissions generated by each experiment are computed based on Equation~\ref{eq:co2}~\citep{strubell-etal-2019-energy,henderson2020towards}.

\begin{equation}\label{eq:co2}
    CO2_emissions = \frac{PUE * kWh * I^{CO2}}{1000}
\end{equation}

where $PUE$, which stands for \textit{Power Usage Effectiveness}, defines how efficiently data centres use energy, i.e. it accounts for the additional energy required to support the compute infrastructure; $kWh$ are the total \textit{kilowatt hours} consumed; $I^{CO2}$ is the CO2 intensity. The $PUE$ and $I^{CO2}$ values vary greatly and depend on a large set of factors. In our computations, similar to~\citep{henderson2020towards} we use averages reported on a global or national level. In particular, we use the global average $PUE$ value reported by~\cite{Ascierto2020_UptimeInstitute_PUE} of $1.59$.\footnote{It is worth noting that the average $PUE$ value follows a descending trend until 2018 when the average $PUE$ is $1.58$ (as reported in the Uptime Institute global data center survey for 2018) and used in the \texttt{experiment-impact-tracker} tool. However, in 2019 the global average $PUE$ is $1.67$, surpassing 2013; in 2020, the value is still high -- $1.59$ (see p. 10 of the 2020 survey~\citep{Ascierto2020_UptimeInstitute_PUE}).}

The $kWh$ are computed as the sum of all power (in watts) drawn per second per GPU which is then divided by $3.600.000$. We ought to note that during training on the P100 workstation and translation (both quantized and not quantized) on both workstations we collected the readings of \verb|nvidia-smi| every second; for the training on the 1080Ti workstation, the readings were collected every five seconds. In order to make the comparison more realistic, for the latter case we interpolated the missing values instead of simply averaging for every 5-second interval.\footnote{We used SciPy's interpolate \url{https://docs.scipy.org/doc/scipy/reference/interpolate.html}}

The carbon intensity is a measure of how much CO2 emissions are produced per kilowatt hour of electricity consumed.\footnote{\url{https://carbonintensity.org.uk/}}. It is a constantly-changing value and as such we consider an average collected over the first half of 2020. The electricityMap project\footnote{\url{https://www.electricitymap.org/}}~\citep{Trangerg2019realtimecarbon} collects and distributes elaborate information related to electricity production and consumption, including carbon intensity per country and per specific timestamp. Since our research is conducted for academic purposes `electricityMap' gave us access to historical data from which we calculated the mean CO2 intensity and standard deviation for both Ireland (IE) and the Netherlands (NL), where our workstations are located (the 1080Ti workstation is located in Ireland, the P100 workstation -- in the Netherlands. The values are as follows: IE: 229.8718 \scalebox{0.75}{$\pm$}77.4026; NL: 399.3685 \scalebox{0.75}{$\pm$}31.9251. 

\section{Power consumption and CO2 footprint}\label{sec:power}
\paragraph{Train and translation times} We first present the run times for each experiment and the training step at which the training was terminated (using early stopping). The run times are shown in Table~\ref{tbl:train_time}. Although we are using the exact same data, software, hyperparameter values\footnote{which are not dependent on the number of GPUs} and random seeds, the training processes on the different hardware deviate due to differences in the hardware, number of GPUs and the NVidia driver.\footnote{\url{https://pytorch.org/docs/stable/notes/randomness.html}}
\begin{table}[ht]
    \centering
    {\setlength\tabcolsep{4.0pt} \begin{tabular}{|c|c|c|c|c|c|c|c|c|c|}\hline
   \multicolumn{2}{|c|}{} &  \multicolumn{3}{c|}{1080Ti}  &  \multicolumn{3}{c|}{P100} \\\cline{3-8}
   \multicolumn{2}{|c|}{System} & Elapsed & \# steps & time/ & Elapsed & \# steps & time/ \\
    \multicolumn{2}{|c|}{} & time (h) & x 1 000 & step & time (h) & x 1 000 & step \\\hline
\multirow{4}{*}{\rotatebox[origin=c]{90}{LSTM}} & EN-FR & 25.08 & 160 & 0.16 & 18.83 & 145 & 0.13\\\cline{2-8}
& EN-ES & 28.41 & 180 & 0.16 & 16.66 & 130 & 0.13\\\cline{2-8}
& FR-EN & 23.51 & 145 & 0.16 & 13.95 & 105 & 0.13\\\cline{2-8}
& ES-EN & 24.38 & 145 & 0.17 & 19.21 & 145 & 0.13\\\hline\hline
\multirow{4}{*}{\rotatebox[origin=c]{90}{TRANS}} & EN-FR & 5.22 & 14.5 & 0.36 & 5.06 & 11 & 0.46\\\cline{2-8}
& EN-ES & 6.6 & 19.5 & 0.34 & 6.06 & 13 & 0.47\\\cline{2-8}
& FR-EN & 6.15 & 17.5 & 0.35 & 4.85 & 11 & 0.44\\\cline{2-8}
& ES-EN & 6.36 & 19 & 0.33 & 6.2 & 13 & 0.48\\\hline
    \end{tabular}}
    \caption{Train time in hours, number of steps and average train time for one step.}
    \label{tbl:train_time}
\end{table}

The values in Table~\ref{tbl:train_time} indicate a larger train time for the LSTM models compared to the the TRANS models, for both the 1080Ti and the more performant P100. We furthermore observe an overall larger train time for the 1080Ti. However, we ought to note that the average time per step for the TRANS models is larger for the P100. We associate these differences with the number of GPUs on which we trained the models -- 4 for the 1080Ti and 3 for the P100. 

In Table~\ref{tbl:translation_time} we present the elapsed time during translation of our test set. Recall that the test set is intentionally large. The translation is conducted on a single GPU and no file is translated by two models at the same time in order to avoid any I/O delays.

\begin{table}[h]
    \centering
    \begin{tabular}{|c|c|c|c|}\hline
   \multicolumn{2}{|c|}{System} &  1080Ti  &  P100 \\\hline
\multirow{4}{*}{\rotatebox[origin=c]{90}{LSTM}} & EN-FR  & 1.52 & 1.84 \\\cline{2-4}
& EN-ES & 1.38 & 1.69 \\\cline{2-4}
& FR-EN & 1.48 & 1.79 \\\cline{2-4}
& ES-EN & 1.34 & 1.62 \\\hline\hline
\multirow{4}{*}{\rotatebox[origin=c]{90}{TRANS}} & EN-FR & 2.63 & 3.01 \\\cline{2-4}
& EN-ES & 2.48 & 2.80 \\\cline{2-4}
& FR-EN & 2.47 & 3.18 \\\cline{2-4}
& ES-EN & 2.45 & 2.69 \\\hline
    \end{tabular}
    \caption{Translation time in hours.}
    \label{tbl:translation_time}
\end{table}

Contrary to what is ought to be expected, from the experiments it can be observed that the translation times on the 1080Ti machine are constantly lower than the ones on the P100.

\paragraph{Power consumption and CO2 emissions.} Below we summarise the measurements collected through the \verb|nvidia-smi| tool and analyse the ecological footprint in terms of CO2 emissions. In Table~\ref{tbl:co2_train} and Table~\ref{tbl:co2_trans} we present the power consumption and carbon dioxide emissions at train and at translation time, respectively.

\begin{table}[h]
    \centering
    {\small \setlength\tabcolsep{2.0pt} 
    \begin{tabular}{|c|c|c|c|c|c||c|c|c|c|}\hline
      \multicolumn{2}{|c|}{} & \multicolumn{4}{c||}{1080Ti} & \multicolumn{4}{c|}{P100}\\\cline{3-10}
     \multicolumn{2}{|c|}{} & Elapsed & Avg. &  & CO2& Elapsed & Avg. &  & CO2\\
     \multicolumn{2}{|c|}{System} & time& power & kWh & (kg) & time & power & kWh & (kg)\\
     \multicolumn{2}{|c|}{} & (h) & (W) & & & (h) & (W) & &\\\hline
\multirow{4}{*}{\rotatebox[origin=c]{90}{LSTM}} & EN-FR & 25.08 & 142.05 & 14.07 & 5.14 \scalebox{0.75}{$\pm$}1.73 & 18.83 & 115.09 & 6.33 & 4.02 \scalebox{0.75}{$\pm$}0.32\\\cline{2-10}
& EN-ES & 28.41 & 140.88 & 15.79 & 5.77 \scalebox{0.75}{$\pm$}1.94 & 16.66 & 113.99 & 5.54 & 3.52 \scalebox{0.75}{$\pm$}0.28\\\cline{2-10}
& FR-EN & 23.51 & 141.85 & 13.15 & 4.81 \scalebox{0.75}{$\pm$}1.62 & 13.95 & 113.48 & 4.63 & 2.94 \scalebox{0.75}{$\pm$}0.24\\\cline{2-10}
& ES-EN & 24.38 & 139.90 & 13.44 & 4.91 \scalebox{0.75}{$\pm$}1.65 & 19.21 & 113.91 & 6.37 & 4.04 \scalebox{0.75}{$\pm$}0.32\\\hline\hline
\multirow{4}{*}{\rotatebox[origin=c]{90}{TRANS}} & EN-FR & 5.22 & 176.70 & 3.64 & 1.33 \scalebox{0.75}{$\pm$}0.45 & 5.06 & 153.47 & 2.27 & 1.44 \scalebox{0.75}{$\pm$}0.12\\\cline{2-10}
& EN-ES & 6.60 & 176.54 & 4.60 & 1.68 \scalebox{0.75}{$\pm$}0.56 & 6.06 & 152.08 & 2.69 & 1.71 \scalebox{0.75}{$\pm$}0.14\\\cline{2-10}
& FR-EN & 6.15 & 176.64 & 4.29 & 1.56 \scalebox{0.75}{$\pm$}0.53 & 4.85 & 151.43 & 2.15 & 1.37 \scalebox{0.75}{$\pm$}0.11\\\cline{2-10}
& ES-EN & 6.36 & 179.48 & 4.50 & 1.64 \scalebox{0.75}{$\pm$}0.55 & 6.20 & 151.59 & 2.74 & 1.74 \scalebox{0.75}{$\pm$}0.14\\\hline
    \end{tabular}}
    \caption{Run-time, power draw and CO2 emissions (kg) at train time. }
    \label{tbl:co2_train}
\end{table}

\begin{table}[ht]
    \centering
   {\small \setlength\tabcolsep{1.8pt} 
    \begin{tabular}{|c|c|c|c|c|c||c|c|c|c|}\hline
      \multicolumn{2}{|c|}{} & \multicolumn{4}{c||}{1080Ti} & \multicolumn{4}{c|}{P100}\\\cline{3-10}
     \multicolumn{2}{|c|}{System} & Elapsed & Average & kWh & CO2& Elapsed & Average & kWh & CO2\\
     \multicolumn{2}{|c|}{} & time (h) & power (w) & & (kg) & time (h) & power (w) & & (kg)\\\hline
\multirow{4}{*}{\rotatebox[origin=c]{90}{LSTM}}  & EN-FR & 1.52 & 157.80 & 0.22 & 0.08 \scalebox{0.75}{$\pm$}0.03 & 1.84 & 90.50 & 0.16 & 0.10 \scalebox{0.75}{$\pm$}0.01\\\cline{2-10}
 & EN-ES & 1.38 & 158.51 & 0.20 & 0.07 \scalebox{0.75}{$\pm$}0.02 & 1.69 & 89.06 & 0.15 & 0.10 \scalebox{0.75}{$\pm$}0.01\\\cline{2-10}
 & FR-EN & 1.34 & 153.43 & 0.19 & 0.07 \scalebox{0.75}{$\pm$}0.02 & 1.79 & 93.14 & 0.16 & 0.10 \scalebox{0.75}{$\pm$}0.01\\\cline{2-10}
 & ES-EN & 1.48 & 154.98 & 0.21 & 0.08 \scalebox{0.75}{$\pm$}0.03 & 1.62 & 89.35 & 0.14 & 0.09 \scalebox{0.75}{$\pm$}0.01\\\hline\hline
\multirow{4}{*}{\rotatebox[origin=c]{90}{TRANS}}  & EN-FR & 2.63 & 188.75 & 0.45 & 0.16 \scalebox{0.75}{$\pm$}0.06 & 3.01 & 104.52 & 0.31 & 0.20 \scalebox{0.75}{$\pm$}0.02\\\cline{2-10}
 & EN-ES & 2.48 & 170.02 & 0.38 & 0.14 \scalebox{0.75}{$\pm$}0.05 & 2.80 & 102.71 & 0.28 & 0.18 \scalebox{0.75}{$\pm$}0.01\\\cline{2-10}
 & FR-EN & 2.47 & 193.34 & 0.47 & 0.17 \scalebox{0.75}{$\pm$}0.06 & 3.18 & 100.93 & 0.31 & 0.20 \scalebox{0.75}{$\pm$}0.02\\\cline{2-10}
 & ES-EN & 2.45 & 175.60 & 0.42 & 0.15 \scalebox{0.75}{$\pm$}0.05 & 2.69 & 104.35 & 0.28 & 0.18 \scalebox{0.75}{$\pm$}0.01\\\hline
    \end{tabular}}
    \caption{Run-time, average power draw and CO2 emissions (kg) at translation time.}
    \label{tbl:co2_trans}
\end{table}

The average power draw, as shown in Table~\ref{tbl:co2_train} and Table~\ref{tbl:co2_trans}, is computed as the sum of all readings (in watt) for all GPUs divided by the number of readings. That is, it is a value that does not take into account the number of GPUs. Comparing the average power draw of LSTM and Transformer (TRANS) models, we observe that for both workstations (both types of GPUs), LSTMs consume less power (at a time) than TRANS models, during both training and translation. However, due to the excessively larger train time of LSTM models, their overall power consumption (reported as kWh) is much larger than that of TRANS models. As such, in our experiments they lead to a larger CO2 emission, as computed by Equation~\ref{eq:co2} for all 4 1080Ti or 3 P100 GPUs: in the case of the 1080Ti, between 2.99 times (4.91 kg to 1.64 kg) for the ES-EN language pair and 3.86 times (5.14 kg to 1.64 kg) for EN-FR; in the case of the P100, between 2.06 times (4.63 kg to 2.15 kg) for FR-EN and 2.79 times (6.33 kb to 2.27 kg) for EN-FR. These results indicate that \emph{for the same data and optimal hyperparameters,\footnote{As recommended in the literature.} LSTMs have a larger ecological footprint at \textbf{train} time.} However, at translation time the observations are reversed -- as suggested by the lower average power consumption for LSTM models and the larger translation time for TRANS models, the CO2 emission regarded to LSTM models is two times lower than that of TRANS models (consistent over all language pairs and GPU types). This would imply that after a certain time of usage, our LSTM models are ``greener'' than the TRANS models. In particular, for the 1080Ti GPUs, our LSTMs would become ``greener'' after 10 to 40 days and for the P100 -- after 9 and 12 days.\footnote{This is the time that would be required for the two types of architectures to consume the same power (in watt) including training (on the same data) and translation. Since LSTM consumes less power at translation time per time unit (e.g. second), any further operation of both LSTM and TRANS models would lead to an overall lower energy consumption by LSTM models.}

When we compare the two types of GPUs, we notice that the P100 is much less power-demanding than the 1080Ti with an average power draw between 113.91 (ES-EN LSTM) and 115.09 (EN-FR LSTM) versus between 139.9 (ES-EN LSTM) and 142.05 (EN-FR LSTM), and between 151.43 (FR-EN TRANS) and 153.47 (EN-FR TRANS) versus between 176.54 (EN-ES TRANS) and 179.48 (FR-EN TRANS).\footnote{At train time.} This is also reflected in the overall power consumption to which a large contributing factor is train time which is much smaller with the P100 workstation. That is, at train time, the power consumption for the P100 GPU workstation is almost three times smaller than that of the 1080Ti machine for LSTM models and around two times smaller for TRANS models. At translation time, while still leading to a lower power consumption, mainly because of the lower average power draw, the differences are much smaller. 

A special attention needs to be paid at the impact of the national carbon intensity. For Ireland it is much lower than that of the Netherlands. These differences have a substantial impact on the carbon emissions -- the TRANS models trained on the P100 workstation in the Netherlands have a larger footprint than in the 1080Ti case in three out of the four cases (EN-FR, EN-ES and ES-EN); at translation time for all models running on the P100 machine, the footprint is larger than in the case of the 1080Ti. 

\begin{figure}[ht]
    \centering
    \includegraphics[width=0.9\textwidth]{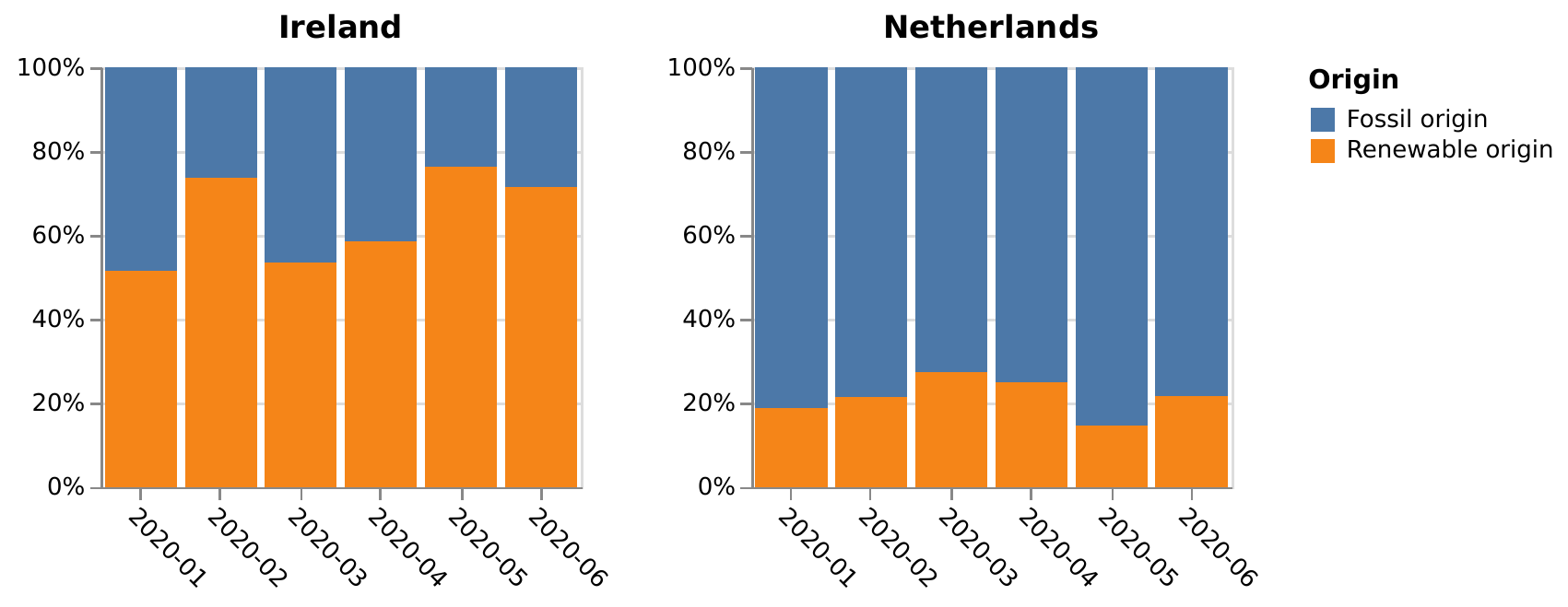}
    \caption{Power origin distribution.}
    \label{fig:power_origin}
\end{figure}

Analysing the data from~\cite{Trangerg2019realtimecarbon} we see that the power for Ireland originates primarily from renewable sources while that is not the case for the Netherlands. Figure~\ref{fig:power_origin} illustrates this for the period between 01/01/2020 and 01/06/2020 on a monthly basis.

\paragraph{The impact} Comparing different types of workstations, different NMT architectures and language pairs as well as the regional factors related to carbon emissions gives us an understanding of the conditions under which a certain amount of electricity is consumed and the factors that we should consider to optimize our NMT-related ecological footprint. To put matters into perspective, we compare our models and the related measurements to common everyday devices. We assess (i) the power draw and (ii) the carbon emissions. For the former, we collect the power consumption as indicated by \emph{Caruna}\footnote{\url{https://www.caruna.fi}}, a Finnish electricity distribution company; for the latter, we gather data from \emph{Carbon Footprint}\footnote{\url{https://www.carbonfootprint.com}}, a company that specialises in carbon emission assessments, LifeCycle Analysis, environmental strategy/planing, etc., and indicates an estimated times or hours of use of common devices. 

\begin{figure}[ht]
\begin{minipage}{.5\textwidth}
    \centering
    \includegraphics[height=0.57\textheight]{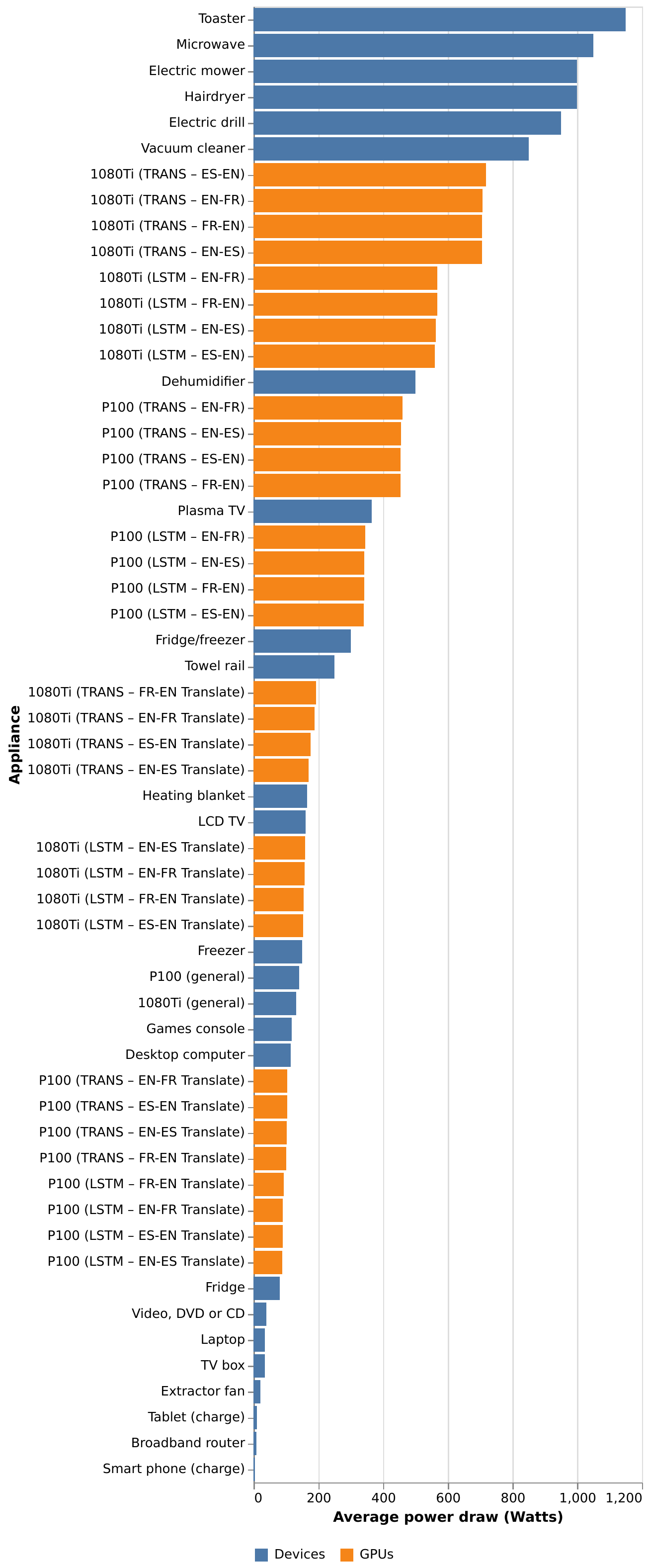}
    \caption{Average power draw.}
    \label{fig:comparison_power}
\end{minipage}
\begin{minipage}{.5\textwidth}
    \centering
    \includegraphics[height=0.55\textheight]{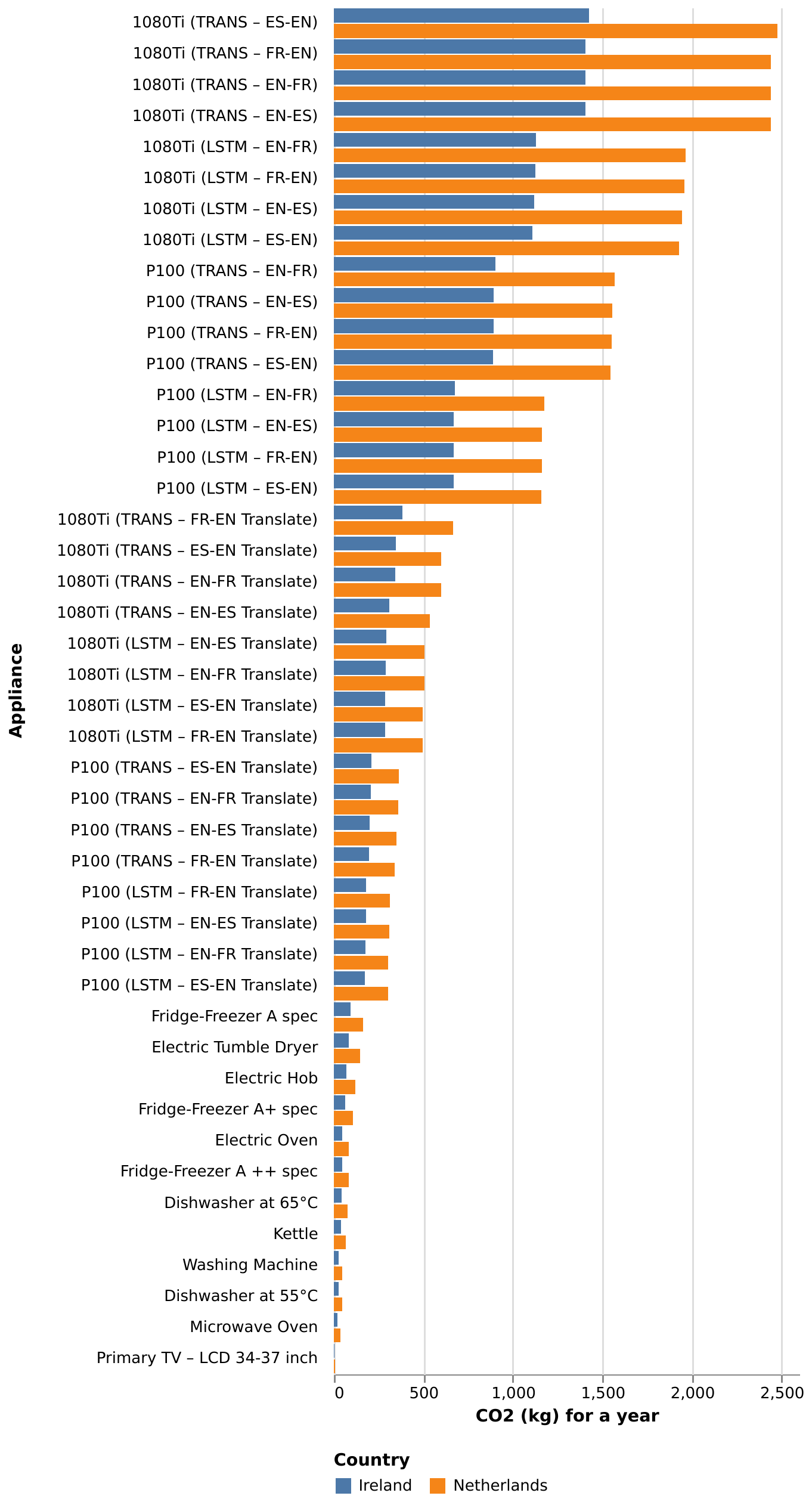}
    \caption{Estimated yearly CO2 emission in kg.}
    \label{fig:comparison_co2}
\end{minipage}
\end{figure}

Figure~\ref{fig:comparison_power} compares the average power draw of the workstations during training and translation for the different models (marked in light blue)\footnote{These are the same values indicated in Table~\ref{tbl:co2_train} and Table~\ref{tbl:co2_trans}.} to common devices (marked in light orange). The common devices are typical household appliances; we have excluded larger, more power demanding devices such as electric shower (8 750 W) since our range, for practical reasons, spans to maximum 1 200 W. From Figure~\ref{fig:comparison_power}, it is clear that the GPUs are less power demanding than some common household appliances. That is, at a given moment in time, a GPU board draws fewer watts of electricity than some common household appliances (e.g. a microwave, an electric mower, etc.). 

However, the usage (or utilisation) time of a GPU is much higher than that of a microwave and as such the power consumption (computed in kWh) as well as the carbon emissions are significantly different. When computing the carbon emissions of a device for a certain period of time, one must take into account the utilisation rate of that device for the given time period, i.e. the number of hours or times the device is used within the considered time interval to the time it is idle. According to~\citep{Nadjaran2017_Renewable} and~\citep{Doyle2020_ROI}, in data centers, the average utilization rate ranges between 20\% and 40\%. However, this is not an ideal scenario for a GPU workstation as the return on investment would be low due to the high costs of GPUs.\footnote{A modern workstation with 3 x Nvidia Tesla V100 costs approximately \EUR 30 000; a workstation with 4 x Nvidia RTX3060 is approximately \EUR 7 000.} As an ideal and rather extreme scenario, we assume that in an industry environment a GPU workstation is utilised 100\%. This, in turn, will indicate the maximum amount of CO2 emissions if GPU workstations of the types we considered are used to train and translate with our models. To compute the impact, in terms of CO2 emissions, we first estimate the carbon emission per unit of time the product of power draw (in watts), utilisation rate\footnote{Values are based on data from \emph{Carbon Footprint}\footnote{\url{https://www.carbonfootprint.com}}} and carbon intensity (for both Ireland and the Netherlands) and second we extrapolate the CO2 emissions over a year based on the recorded values for the given time period. Our estimates are shown in Figure~\ref{fig:comparison_co2}. 

Here, however, we ought to note that while at train time each workstation utilizes all of its GPUs, that is not the case for the translation. As such, we advise the reader to consider these as two different types of measurements -- one where all GPUs are utilized (train time) and one where only one GPU is used (translation time). 

From Figure~\ref{fig:comparison_co2} we observe that in one year, at 100\% utilization, a GPU workstation that is used to train simple models, such as ours, could produce up to $2.500$ kg of CO2.\footnote{One can also multiply the CO2 emissions at translation time by 4 for the 1080Ti or by 3 for the P100 GPUs and get an indication about how much CO2 emissions would be generated if a workstation (with 4 or 3 GPUs, as in our case) is utilized 100\% at translation time. That is, either all GPUs translate using the same model in parallel, or different models are used at the same time for different translation jobs.} This is approximately equal to the CO2 emissions from the electricity consumption of two small households in the UK.\footnote{According to data from \url{https://www.carbonindependent.org}.}

\section{Optimizing at inference time through model quantization}\label{sec:mitigation}
In an academic environment, the main use of MT is as an experimentation ground for research. Thus, it is often the case that many NMT models are trained and evaluated on a small test set. However, within an industry environment, the main purpose of an NMT model is to be used to translate as much new content as possible before updating or rebuilding it. From the evaluation summarised in Section~\ref{sec:power} we see that, at translation time the P100 workstation uses less energy than a 1080Ti workstation (see Table~\ref{tbl:co2_trans}); furthermore at translation time the P100 workstation uses less energy than both workstations at train time (see Table~\ref{tbl:co2_train}). And while this is an indication that investing in a higher-end GPU workstation is better (in terms of electricity consumption and carbon emissions) we ought to investigate other means to optimize both power consumption and inference time. With that regards, in this section we investigate how quantizing the model can improve both energy consumption, inference time as well as what the impact is on the translation quality.

\paragraph{Quantization} Deep neural networks contain a huge amount of parameters (biases, weights) that are adapted during training to reduce the loss. These are typically stored as 32-bit floating point numbers. In every forward pass all these parameters are involved in computing the output of the network. The large precision requires more memory and processing power than, e.g. integer numbers. Quantization is the process of approximating a neural network's parameters by reducing their precision. A quantized model executes some or all of the operations on tensors with integers rather than floating point values.\footnote{\url{https://pytorch.org/docs/stable/quantization.html}} Quantization is a term that encapsulates a broad range of approaches to the aforementioned process: binary quantization~\citep{Courbariaux2016_binarynet}, ternary~\citep{Lin2015_ternary_quantization,Li2016_ternary_quantization}, uniform~\citep{Jacob2018_uniform_quantization} and learned~\citep{Zhang2018_learned_quantization}, to mention a few. The benefits of quantization are a reduced model size and the option to use more high performance vectorized operations, as well as, the efficient use of other hardware platforms\footnote{For example, the $2^{nd}$ generation Intel$^{\tiny\circledR}$ Xeon$^{\tiny\circledR}$ Scale processors incorporate an INT8 data type acceleration (Intel$^{\tiny\circledR}$ DL Boost Vector Neural Network Instructions (VNNI)~\citep{Fomenko2018_vnni}), specifically designed to accelerate neural network-related computations~\citep{Rodriguez2018_lower}.}. However, more efficient quantized models may suffer from worse inference quality. In the field of NMT~\cite{Bhandare2019_quantization} quantize trained Transformer models to a lower precision (8-bit integers) for inference on Intel$^{\tiny\circledR}$ CPUs. They investigate three approaches to quantize the weights of a Transformer model and achieve a drop in performance of only 0.35 to 0.421 BLEU points (from 27.68 originally). \cite{Prato2020_quantization} investigate a uniform quantization for Transfromer, quantizing matrix multiplications and divisions (if both the numerator and denominator are second or higher rank tensors), i.e. all operations that could improve the inference speed. Their 6-bit quantized EN-DE Transformer base model is more than 5 times smaller than the baseline and achieves higher BLEU scores. For EN-FR, the 8-bit quantized model which achieves the highest performance is almost 4 times smaller than the baseline. With the exception of the 4-bit fully quantized models and the naive approach, all the rest show a significant reduction in model size with almost no loss in translation quality (in terms of BLEU).

The promising results from the aforementioned works motivated us to investigate the power consumption of quantized versions of our models running on a GPU. We used the \verb|CTranslate2| tool of OpenNMT. \footnote{\url{https://github.com/OpenNMT/CTranslate2}} As noted in the repository ``CTranslate2 is a fast inference engine for OpenNMT-py and OpenNMT-tf models supporting both CPU and GPU execution. The goal is to provide comprehensive inference features and be the most efficient and cost-effective solution to deploy standard neural machine translation systems such as Transformer models.''

\paragraph{Quality of quantized Transformer models} We quantized our Transformer models to INT8 and INT16 and translated our test sets on the 1080Ti and P100 workstations. On each workstation, we quantized the models that were originally trained on that machine. After translation, we scored BLEU and TER the same way as with our normal models. The quality results are summarised in Table~\ref{tbl:quantized_results}.

\begin{table}[ht]
    \centering
    {\small \setlength\tabcolsep{1.8pt} 
    \begin{tabular}{|c|l|c|c|c|c|c|c|c|c|}\hline
         & & \multicolumn{2}{c|}{EN-FR} & \multicolumn{2}{c|}{EN-ES} & \multicolumn{2}{c|}{FR-EN} & \multicolumn{2}{c|}{ES-EN}\\\cline{2-10}
         GPU & Prec. & BLEU$\uparrow$ & TER$\downarrow$ & BLEU$\uparrow$ & TER$\downarrow$& BLEU$\uparrow$ & TER$\downarrow$ & BLEU$\uparrow$ & TER$\downarrow$ \\\hline
 \multirow{3}{*}{\rotatebox[origin=c]{90}{1080Ti}} & {\scriptsize FP32} & \textbf{37.2} & \textbf{48.7} & \textbf{40.9} & \textbf{43.4} & \textbf{37.0} & \textbf{46.4} & \textbf{41.3} & \textbf{41.4}\\\cline{2-10}
 & {\scriptsize INT16} & 36.4 & 49.5 & 40.6 & 44.1 & 36.6 & 47.3 & 40.9 & 44.0 \\\cline{2-10}
 & {\scriptsize INT8} & 36.3 & 49.5 & 40.5 & 44.1 & 36.6 & 47.3 & 40.8 & 44.1 \\\hline\hline

 \multirow{3}{*}{\rotatebox[origin=c]{90}{P100}} & FP32 & \textbf{37.4} & \textbf{48.4} & \textbf{40.9} & \textbf{43.3} & \textbf{37.3} & \textbf{47} & \textbf{41.6} & \textbf{42.5}\\\cline{2-10}
 & {\scriptsize INT16} & 36.4 & 49.1 & 40.7 & 44.0 & 36.5 & 50.6 & 40.5 & 43.8 \\\cline{2-10}
 & {\scriptsize INT8} & 36.4 & 49.2 & 40.7 & 44.0 & 36.5 & 50.4 & 40.5 & 43.8 \\\hline
    \end{tabular}}
    \caption{Evaluation metrics for quantized and baseline Transformer models. FP32 stands for 32-bit floating point; INT16 or INT8 stand for 16-bit or 8-bit integer. FP32 is the default, non-quantised model (See Table~\ref{tbl:mt_eval}).}
    \label{tbl:quantized_results}
\end{table}

\paragraph{Energy considerations for quantized Transformer models}

We measured the power draw during the translation process with our quantized models and computed the consumed kWh as well as CO2 emissions per model and per region. Our results are summarised in Table~\ref{tbl:co2_trans_quant}.

\begin{table}[h]
    \centering
    {\small \setlength\tabcolsep{2pt} 
    \begin{tabular}{|ll|c|c|c|c||c|c|c|c|}\hline
    \multicolumn{2}{|c|}{} & \multicolumn{4}{c||}{1080Ti} & \multicolumn{4}{c|}{P100}\\\cline{3-10}
     \multicolumn{2}{|c|}{} & Elapsed & Avg. &  & CO2& Elapsed & Avg. &  & CO2\\
     \multicolumn{2}{|c|}{System} & time& power & kWh & (kg) & time & power & kWh & (kg)\\
     \multicolumn{2}{|c|}{} & (h) & (W) & & & (h) & (W) & &\\\hline
{\scriptsize INT16} & EN-FR & 5.17 & 130.61 & 0.13 & 0.05 $\pm$ 0.02 & 0.79 & 81.54 & 0.01 & 0.01 $\pm$0.00\\\hline
{\scriptsize INT8} & EN-FR & 4.66 & 115.99 & 0.11 & 0.04 $\pm$ 0.01 & 2.16 & 49.06 & 0.02 & 0.01 $\pm$0.00\\\hline\hline
{\scriptsize INT16} & EN-ES & 4.45 & 158.96 & 0.14 & 0.05 $\pm$ 0.02 & 0.99 & 65.4 & 0.01 & 0.01 $\pm$0.00\\\hline
{\scriptsize INT8} & EN-ES & 4.15 & 124.40 & 0.10 & 0.04 $\pm$ 0.01 & 1.00 & 68.01 & 0.01 & 0.01 $\pm$0.00\\\hline\hline
{\scriptsize INT16} & FR-EN & 4.57 & 139.38 & 0.13 & 0.05 $\pm$ 0.02 & 1.28 & 67.57 & 0.02 & 0.01 $\pm$0.00\\\hline
{\scriptsize INT8} & FR-EN & 4.39 & 107.87 & 0.09 & 0.03 $\pm$ 0.01 & 1.29 & 68.39 & 0.02 & 0.01 $\pm$0.00\\\hline\hline
{\scriptsize INT16} & ES-EN & 4.45 & 131.66 & 0.12 & 0.04 $\pm$ 0.01 & 1.02 & 68.33 & 0.01 & 0.01 $\pm$0.00\\\hline
{\scriptsize INT8} & ES-EN & 4.03 & 117.48 & 0.09 & 0.03 $\pm$ 0.01 & 1.04 & 67.25 & 0.01 & 0.01 $\pm$0.00\\\hline
\end{tabular}}
    \caption{Run-time, average power draw and CO2 emissions (kg) at translation time for quantized models.}
    \label{tbl:co2_trans_quant}
\end{table}

Comparing these results to the ones in Table~\ref{tbl:co2_trans} (TRANS) we first notice the increased translation time for the quantized models running on the 1080Ti machine -- from 2.45 to 4.03 (INT8) and 4.45 (INT16) hours for the fastest ES-EN and from 2.63 to 5.17 (INT16) and 4.66 (INT8) for the slowest EN-FR. However, due to the lower power draw the overall energy consumption as well as the CO2 emissions at translation time with these models (on the 1080Ti workstation) is still lower than for the non-quantized models (on the same workstation). 

When comparing the performance of these models on the P100 workstation we notice a much lower translation time over all models, even if the difference between the EN-FR / INT16 model and the non-quantized EN-FR Transformer model is not so drastic. At the same time the power draw for all models is lower than for their non-quantized version, leading to a very low electricity consumption and low carbon emissions.  

We ought to note that the time for quantization, i.e. the process of converting a non-quantized Transformer model into a quantized one is very low (between 6 and 12 seconds on the P100 workstation). Furthermore, quantization is very much suitable for inference on CPU. Based on the results in Table~\ref{tbl:quantized_results} and Table~\ref{tbl:co2_trans_quant}, the low quantization time and the fact that quantized models can easily be run on CPU, we would recommend quantized models at translation time for large-scale translation projects in the pursuit of greener MT.

\section{Conclusions and future work}
In the era of deep learning, neural models are continuously pushing the boundaries in NLP, MT including. The ever growing volumes of data and the advanced, larger models keep delivering new state-of-the-art results. A facilitator for these results are the innovations in general purpose GPU computing, as well as in the hardware itself, i.e. GPUs. The embarrassingly parallel processing required for deep learning models is easily distributed on the thousands of processing cores on a GPUs, making training and inference with such models much more efficient than on CPUs. However, GPUs are much more power demanding and as such have a higher environmental impact. In this chapter we discussed considerations related to the power consumption and ecological footprint, in terms of carbon emissions associated with the training and the inference with MT models. 

After briefly presenting the evolution of MT and the shift to GPUs as the core processing technology of (N)MT, we discuss the related work addressing the issues of power consumption and environmental footprint of computational models. We acknowledge that the work of colleague researchers and practitioners, such as~\cite{strubell-etal-2019-energy} and~\cite{Schwartz2020GreenA}, among others, raises awareness about the environmental footprint that deep learning models have, and we would like to join them in their appeal towards ``greener'' AI. This could be achieved through optimizing models through quantization, as discussed in Section~\ref{sec:mitigation}, but also, through reusability, smarter data selection, knowledge distillation and others.  

To outline the realistic dimensions of power consumption and environmental footprint of NMT, we analysed a number of NMT models, running (training or inference) on two types of GPUs -- a consumer GPU card (NVidia GTX 1080Ti), designed to work on a desktop machine and a workstation GPU (NVidia Tesla P100) developed for heavy loads of graphics or neural computing. We reported results for training both LSTM and Transformer models for English-French and English-Spanish (and vice-versa) language pairs on data from the Europarl corpus. These models were trained on approximately 1.5M parallel sentences and used to translate large test sets that include approximately 500.000 sentences each. Our results and analysis show that, while a Transformer model is much faster and as such much more power efficient than an LSTM at train time, at translation time Transformer models are lagging behind LSTM models, in terms of power consumption, speed as well as carbon emissions. We also note that using the more expensive P100 is preferable in almost every case. An exception is the slightly higher translation time, which however, comes with the benefit of a largely reduced power consumption. 

Additionally, we also note the impact of electricity sources on the carbon emissions by investigating two different countries, each of which has a different distribution between fossil and renewable energy sources -- Ireland with a larger portion of renewable energy and the Netherlands with a larger portion of fossil sources. 

Together with the aforementioned contributions, we also aim to motivate researchers to devote time, efforts and investments in developing more ecological solutions. We are already looking into model reusability, data selection and filtering, multi-objective optimization of hyperparameters and other approaches that reduce the environmental footprint of NMT. 

\section*{Carbon Impact Statement }
This  work  contributed  50.77 $\pm$11.37kg  of CO2eq to  the  atmosphere  and  used  111.55  kWh  of  electricity.

\section*{Acknowledgements} We would like to thank \url{electicityMap.org} for their responsiveness to our queries and providing us with valuable information.

\bibliography{bibliography}
\bibliographystyle{acl_natbib}
\end{document}